\documentclass[]{interact}

\usepackage{color}
\usepackage[dvipsnames]{xcolor}

\usepackage{epstopdf}
\usepackage[caption=false]{subfig}

\usepackage[numbers,sort&compress]{natbib}
\bibpunct[, ]{[}{]}{,}{n}{,}{,}
\renewcommand\bibfont{\fontsize{10}{12}\selectfont}
\makeatletter
\def\NAT@def@citea{\def\@citea{\NAT@separator}}
\makeatother

\theoremstyle{plain}

\theoremstyle{definition}

\theoremstyle{remark}

\usepackage{array}
\usepackage{enumitem}
\usepackage{adjustbox}

\begin{document}

\title{Nearest Neighbor Future Captioning: Generating Descriptions for Possible Collisions in Object Placement Tasks}

\author{
\name{Takumi Komatsu\textsuperscript{a}\thanks{CONTACT Takumi Komatsu. Email: tak3k\_1999@keio.jp}\thanks{This article has been accepted for publication in Advanced Robotics, published by Taylor Francis.} and Motonari Kambara\textsuperscript{a} and Shumpei Hatanaka\textsuperscript{a} and Haruka Matsuo\textsuperscript{a} and Tsubasa Hirakawa\textsuperscript{b} and Takayoshi Yamashita\textsuperscript{b} and Hironobu Fujiyoshi\textsuperscript{b} and Komei Sugiura\textsuperscript{a}}
\affil{\textsuperscript{a}Keio University, 3-14-1 Hiyoshi, Kohoku, Yokohama, Kanagawa 223-8522, Japan; \textsuperscript{b}Chubu University, 1200 Matsumotocho, Kasugai, Aichi 487-8501, Japan.}
}

\maketitle

\begin{abstract}
Domestic service robots (DSRs) that support people in everyday environments have been widely investigated. However, their ability to predict and describe future risks resulting from their own actions remains insufficient.
In this study, we focus on the linguistic explainability of DSRs.
Most existing methods do not explicitly model the region of possible collisions; thus, they do not properly generate descriptions of these regions.
In this paper, we propose the Nearest Neighbor Future Captioning Model that introduces the Nearest Neighbor Language Model for future captioning of possible collisions, which enhances the model output with a nearest neighbors retrieval mechanism.
Furthermore, we introduce the Collision Attention Module that attends regions of possible collisions, which enables our model to generate descriptions that adequately reflect the objects associated with possible collisions.
To validate our method, we constructed a new dataset containing samples of collisions that can occur when a DSR places an object in a simulation environment.
The experimental results demonstrated that our method outperformed baseline methods, based on the standard metrics.
In particular, on CIDEr-D, the baseline method obtained 25.09 points, whereas our method obtained 33.08 points.
\end{abstract}


\section{Introduction
\label{intro}
}

In aging societies, the shortage of home care workers has become a serious social problem. Domestic Service Robots (DSRs) are one of the possible solutions \cite{yamamoto2019development}.
Delivering everyday objects in cluttered environments is a critical task for DSRs.
Notably, when DSRs manipulate objects, there is a risk of collisions with other objects, resulting in damage to the DSRs or objects.
It would be useful to predict the possible collisions and then explain them to the user in natural language. However, such functionality remains insufficient.

In this study, we focus on generating descriptions of possible collisions when DSRs place objects. 
It is desirable to generate captions describing subsequent events triggered by the initial contact.

\begin{figure}[t]
    \centering
    \includegraphics[width=\linewidth]{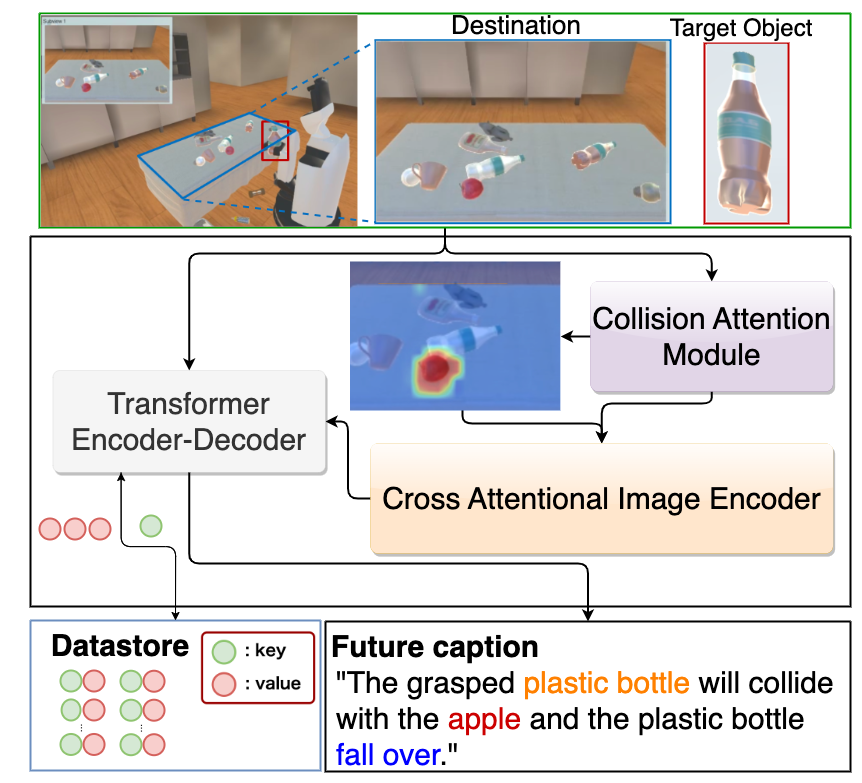}
    \caption{\normalsize Overview of our method: Given an image of the current destination and the target object, our method generates descriptions of possible collisions.}
    \label{fig:eye_catch}
    \vspace{-5mm}
\end{figure}

For example, when a DSR is requested to place a plastic bottle on a desk and it appears that the plastic bottle may collide with an apple, it is desirable to inform the user before taking the action, using an explanation such as ``The grasped plastic bottle will collide with an apple and the plastic bottle fall over'' as shown Fig.~\ref{fig:eye_catch}.
This task is challenging because it requires the DSRs to predict possible collisions and provide explanations.
For instance, template matching, which simply fills predefined structures with recognized object names, fails to account for the subsequent events following the initial contact. 
Consider a case where even if objects involved in collisions, such as a plastic bottle and an apple, are correctly identified, a fixed template cannot appropriately describe the subsequent results, such as ``the plastic bottle will fall over.'' 
To appropriately describe such events, models must be capable of generating various expressions on a case-by-case basis.
Therefore, this task requires a free-form text generation model, rather than relying on template matching.

Although several previous studies have explored sentence generation for possible collisions, existing methods do not explicitly model the region of possible collisions, resulting in low-quality descriptions \cite{yi2019clevrer, kambara2022relational, chen2022comphy}.

In this paper, we propose the Nearest Neighbor Future Captioning Model (NNFCM) that introduces the Nearest Neighbor Language Model (NNLM) \cite{khandelwal2019generalization} for future captioning of possible collisions, which enhances the model output by employing a nearest neighbors retrieval mechanism.
Furthermore, we introduce the Collision Attention Module that attends regions of possible collisions, which enables our model to generate descriptions that adequately reflect the objects associated with possible collisions. Our code is available at this URL\footnote{https://github.com/keio-smilab24/RelationalFutureCaptioningModel.git}.

Our method differs from existing methods in that it introduces the NNLM, which has not been fully utilized in vision and language studies including future captioning. 
The NNLM has substantially outperformed the transformer-based models in the field of language modeling \cite{khandelwal2019generalization} and machine translation \cite{khandelwal2020nearest}. 
The NNLM also predicts tokens by applying a nearest neighbor classifier over a large Datastore of cached examples, which is expected to generate diverse expressions.
This capability is crucial because a minor collision can lead to more serious events in object placement tasks.
For example, in the case shown in Fig.~\ref{fig:eye_catch}, even if the collision between the apple and the plastic bottle is minor, it can lead to a more significant event, such as the bottle falling off the desk.
Given these considerations, the model needs to generate descriptions that appropriately represent the range of possible subsequent events.
Furthermore, our Collision Attention Module extracts the attention regions of possible collisions, which is expected to generate descriptions that appropriately reflect the regions of possible collisions. 

The main novelties of this paper are summarized as follows:
\begin{itemize}
 \item We propose the NNFCM that introduces the NNLM for future captioning, specifically regarding possible collisions. This is one of the first attempts to introduce the NNLM into multimodal language generation. 
 \item We introduce the Collision Attention Module, which extracts attention regions of possible collisions.
 \item We introduce the Cross Attentional Image Encoder, which models the relationship between the target object and destination.
\end{itemize}

\begin{figure}[t]
    \centering
    \includegraphics[clip, height=40mm]{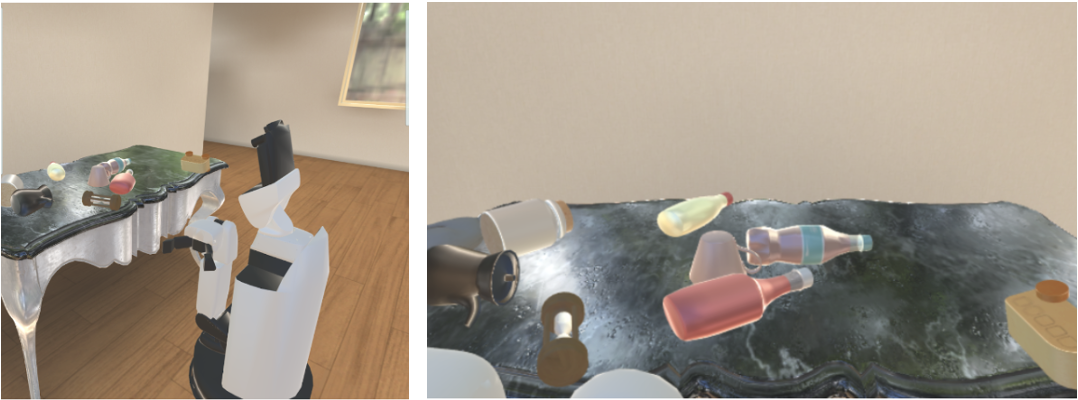}
    \caption{\normalsize A typical scene of the task. Left: an image of the simulation environment. Right: an image of the destination. }
    \label{fig:scene}
    \vspace{-5mm}
\end{figure}

\vspace{-2.0mm}
\section{Related Work
\label{related}
}
\vspace{-0.8mm}

There have been many studies in the field of image captioning 
\cite{xu2015show, krishna2017dense, wang2018video, lei2020mart}.
The field is roughly divided into static image captioning and video captioning.
\cite{stefanini2022show} is one of the review papers in the field of image captioning that provides a comprehensive overview and categorization of approaches.
\cite{aafaq2019video} is another review paper that provides a comprehensive summary of each method, standard dataset, and standard metric in video captioning. Caption generation task using history information can be divided into subtasks, such as future captioning tasks and video captioning tasks.

The future captioning task \cite{hosseinzadeh2021video} generally aims to generate descriptions of an event that will occur in the future based on past and current information.
There are several conventional methods such as \cite{hosseinzadeh2021video, mori2021, mahmud2021103230}.
For example, \cite{mori2021} generates descriptions of events in the near future using in-vehicle camera images and vehicle motion information from the past to the present. The video-and-language event prediction task \cite{lei2020mart} and video prediction task \cite{deng2021sketch} are similar to the future captioning task in that they aim to predict events in the future.
However, they differ from the future captioning task in that the prediction targets are events and videos.

Numerous studies have been conducted in video captioning (e.g. \cite{wang2018video, lei2020mart, zhang2021open, zhao2021multi, sun2019videobert, luo2020univl}). Recent studies have demonstrated the effectiveness of large-scale pre-training models in the field \cite{sun2019videobert, luo2020univl, li2020hero}.
\cite{sun2019videobert, luo2020univl} are typical large-scale pre-training models.
Moreover, HERO \cite{li2020hero} outperforms existing methods in various tasks, including retrieval tasks.
Previous reports have also investigated other captioning tasks in robotics \cite{ogura2020alleviating, magassouba2020multimodal}. 
\cite{ogura2020alleviating} investigated the fetching instruction generation task, which is one of the image captioning tasks. This task aims to generate an instruction for moving the specified object to the destination in a given scene. Specifically, ABEN \cite{ogura2020alleviating} and Multi-ABN \cite{magassouba2020multimodal} generate the region of attention for the generated instructions using ABN \cite{fukui2019attention}.

In the field of image captioning, there are several standard datasets used to compare the methods.
For example, Flickr30K \cite{plummer2015flickr30k} consists of pictures that capture people engaged in daily activities and events. The dataset contains over 31,000 images collected from the Flickr website, and each image has five reference descriptions provided by human annotators.
In addition, Microsoft COCO \cite{lin2014microsoft} is a large-scale object detection, segmentation, and captioning dataset. It consists of images of complex daily scenes with common objects in their natural context. It contains more than 120,000 images, and each is annotated with five reference descriptions.
Additionally, in the field of video captioning, there are several standard datasets used to compare the methods.
For example, the YouCook2 dataset \cite{zhou2018towards} contains cooking videos. It consists of 2,000 videos on 89 recipes. Each video is annotated with the start and end time of each procedure and a reference description.
In addition, the ActivityNet dataset \cite{heilbron2015activitynet} is aimed at understanding human behavior. It contains 137 videos, and the behaviors in the videos are categorized into 203 classes.

Some papers provide a comprehensive review of robot collision and describe the methods used to ensure safe interactions in  shared environments with humans.
For example, machine learning has been used for developing post-collision decision-making strategies \cite{haddadin2017robot}.
Additionally, the ThreeDWorld platform offers a multi-agent simulation environment where physical interactions and multimodal sensory data integration are realized, enhancing the agents' ability to understand and navigate complex 3D settings \cite{gan2021threedworld}.
In contrast, our research focuses on generating descriptions of potential collisions and subsequent events before the robot executes an action.
This emphasizes an interactive approach based on predictions with users. 
There are also some real-time detection methods based on sensory information using reactive control strategies \cite{haddadin2008collision} or DNNs \cite{heo2019collision}.
Furthermore, several methods predict collisions using an image and a placement policy \cite{mottaghi2016happens, magassouba2021predicting}.
Our method differs from these methods in that we focus on describing future collisions and events based on visual information.
This is useful because if the model can describe the possible collisions and subsequent events in natural language, users can easily decide whether the robot should act or not. This can lead to improved safety for the robots and the environment.

Our method is inspired by the collision-related future captioning model such as RFCM \cite{kambara2022relational}.
Unlike RFCM, our model takes images of the destination before task execution as input and the novelty of our work is summarized as follows:
\begin{itemize}
 \item We propose an NNFCM that introduces the NNLM for future captioning, specifically regarding possible collisions. 
 \item We introduce the Collision Attention Module, which extracts attentional regions of possible collisions.
 \item We introduce the Cross Attentional Image Encoder, which models the relationship between the target object and destination.
\end{itemize}

\begin{figure*}
    \centering
    \includegraphics[clip,width=\linewidth,]{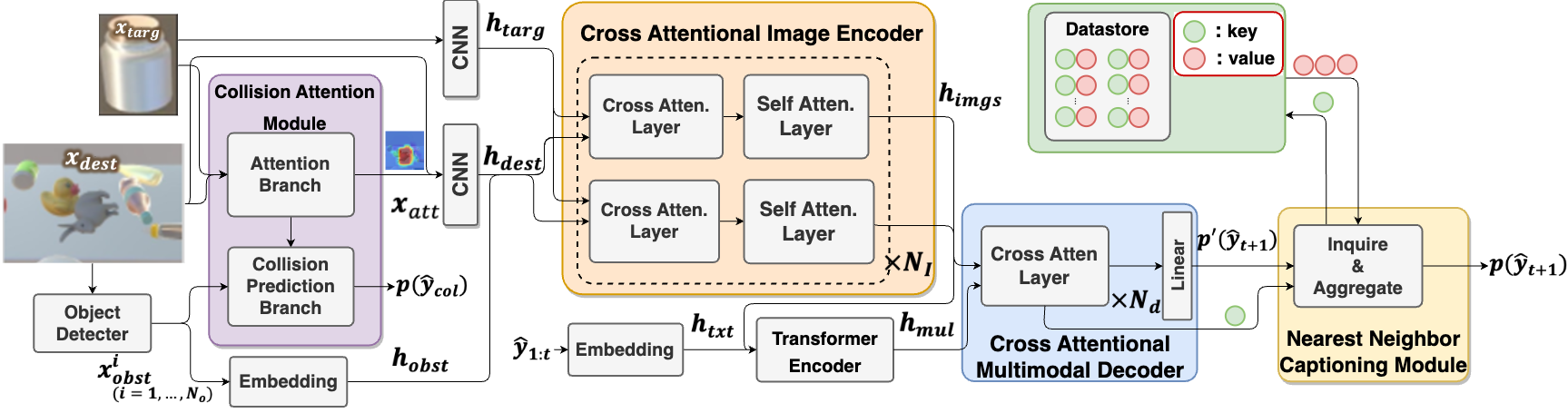}
    \vspace{-5mm}
    \caption{\normalsize The framework of our model. The Cross Attentional Image Encoder consists of two stacked blocks, which enable the model to understand the mutual relationship between the target object and its destination. ``Cross Atten. Layer'' and ``Self Atten. Layer'' denote Cross Attention Layer and Self Attention Layer, respectively. }
    \label{fig:model}
    \vspace{-5mm}
\end{figure*}

\vspace{-1.0mm}
\section{Problem Statement
\label{sec:problem}
}
\vspace{-0.8mm}


The target task of this study is collision-related future captioning.
This involves the predictive task of generating a description of the future situation from an image before motion execution. In this task, it is desirable to output an appropriate description of likely collisions and subsequent events following the initial contact. Fig.~\ref{fig:scene} shows a typical scene of the task, where the left and right images show the simulation environment and destination, respectively. In this scene, the ground truth is ``The gripped plastic bottle collided with the ketchup container because it was being placed on top of the ketchup container.''

The task is characterized by the following:
\begin{itemize}
    \item \textbf{Input:} An RGBD image of the destination, and an RGBD image of a target object.
    \item \textbf{Output:} A sentence describing a possible collision.
\end{itemize}

We define the terms used in this paper as follows:
\begin{itemize}
    \item Destination: A region where the robot places objects, (e.g., a shelf or desk).
    \item Target object: An everyday object to be placed, (e.g., a plastic bottle or can).
    \item Obstacles: Objects already placed on a destination.
\end{itemize}

In this study, we assume that the obstacles are unknown because it significantly reduces the applicability of this study if we assume that the obstacles are known. Indeed, in a domestic environment where everyday objects are cluttered on a desk or table, it is unlikely that these objects are known beforehand and their 3D models are available. Therefore, in most cases, it is not possible to simply calculate the size of the obstacles.
In addition, the input is an image, not a video. We do not use video because of the significant problems with applicability. Using video up to a certain time $t$ requires the assumption that no collisions occur until that time. This is a limitation that significantly reduces the applicability of our study. In practice, collisions are caused by multiple factors, including the trajectory and the size of the objects. This makes it difficult to appropriately predict when a collision may occur. In particular, the size of the obstacles is unknown, as explained above. Therefore, it is preferable to predict collisions based on an image and other information related to actions or motions. 

We utilize typical automatic evaluation metrics: BLEU-4 \cite{papineni2002bleu}, METEOR \cite{banerjee2005meteor}, ROUGE-L \cite{lin2004rouge} and CIDEr-D \cite{vedantam2015cider}. We also use a simulation environment instead of a physical robot because we deal with collisions related to robots. If we use a physical robot, there may be a risk of the robot malfunction or causing damage to the environment and objects.

\vspace{-0.5mm}
\section{Proposed Method
\label{method}
}

In this paper, we propose Nearest Neighbor Future Captioning Model (NNFCM). Our method introduces the
NNLM \cite{khandelwal2019generalization}, which has been shown to substantially outperform the transformer-based models in machine translation \cite{khandelwal2020nearest}. To the best of our knowledge, this is one of the first attempts to introduce the NNLM to multimodal language generation. Although we apply it to future captioning in this paper, it can be broadly applicable to other captioning tasks as well.

The novelties of the proposed method are as follows:
\begin{itemize}
 \item We introduce the NNLM \cite{khandelwal2019generalization} for future captioning.
 \item Unlike the RFCM \cite{kambara2022relational}, we introduce the Collision Attention Module to focus on the attentional regions related to collisions.
 \item We introduce the Cross Attentional Image Encoder to model the relationship between the target object and destination.
\end{itemize}

Fig.~\ref{fig:model} shows the structure of the proposed method. It consists of four main modules: Collision Attention Module, Cross Attentional Image Encoder, Cross Attentional Multimodal Decoder and Nearest Neighbor Captioning Module.

\subsection{Input}

The input to our model is defined as follows:
\begin{align*}
        \{\bm{x}_\mathrm{dest}, \bm{x}_\mathrm{targ}, \bm{x}^{i}_\mathrm{obst}|i=1,...N_o\},
\end{align*}
where, $\bm{x}_\mathrm{dest} \in \mathbb{R}^{c\times h \times w}$, $\bm{x}_\mathrm{targ} \in \mathbb{R}^{c\times h \times w}$, $\bm{x}^{i}_\mathrm{obst} \in \mathbb{R}^{1024}$ denote the RGBD image of the destination, the RGBD image of the target object, and the feature vector of the region of the $i$-th obstacle, respectively. In addition, $N_o$ denotes the number of detected obstacles.

Faster R-CNN \cite{ren2015faster} is used for extracting $\bm{x}^{i}_\mathrm{obst}$ from $\bm{x}_\mathrm{dest}$. 
Here, the output of the FC-6 layer of the RoI Pooling layer consisting of ResNet50 \cite{he2016deep} is used. 
We use a 7-dimensional vector $\{x_1, y_1, x_2, y_2, w, h, w \times h \}$ as the positional encoding for the visual features of the regions.
Here, $(x_1, y_1)$ and $(x_2, y_2)$ denote the coordinates of the upper-left and lower-right vertices of each region, respectively. In addition, $w$ and $h$ denote the width and height, respectively. 
Then, we obtain $\bm{h}^{i}_\mathrm{obst}$ using a fully-connected layer and a normalization layer.
$\bm{x}_\mathrm{targ}$ which is resized to $224 \times 224$ and standardized is input into CNN module to obtain $\bm{h}_\mathrm{targ}$. The proposed method is applicable to physical robots using the same inputs as in the simulation environment.

\subsection{Collision Attention Module (CAM)}
This module extracts attention regions of possible collisions. The module is an extended version of the method proposed in \cite{magassouba2021predicting}. It consists of the Collision Prediction Branch and Attention Branch. The Collision Prediction Branch predicts a collision when placing a target object and the Attention Branch generates an attention map for possible collisions. Here, the attention map represents the importance of each pixel. The CAM predicts the collision probability when the robot places objects according to the policy. Therefore, the attention map generated by the CAM will should focus on the object placement positions in each episode. In other words, CAM outputs an attention map that considers the policy. The details are explained in \cite{magassouba2021predicting}. The input of this module is $\{\bm{x}_\mathrm{dest}, \bm{x}_\mathrm{targ}, \bm{x}^{i}_\mathrm{obst}|i=1,...N_o\}$ and the output is $p(\hat{y}_{col})$ and an attention map $\bm{x}_\mathrm{att}$, where $\hat{y}$ denotes a binary label indicating the presence or absence of a collision. $\bm{x}_\mathrm{dest}$ and $\bm{x}_\mathrm{att}$ are resized to $224 \times 224$ and standardized. Then $\{\bm{x}_\mathrm{dest}, \bm{x}_\mathrm{att}\}$ are input into CNN module to obtain $\bm{h}_{dest}$.

\subsection{Cross Attentional Image Encoder (CAIE)}
This encoder models the relationship between the target object and the destination. The encoder consists of $N_I$ layers. Each layer consists of a Cross-Attention layer, a Multi-Head Attention layer and a Feed-Forward Network layer as in the Transformer \cite{vaswani2017attention}.
We define cross-attention operations using arbitrary matrices $X_A$ and $X_B$, as follows:
\begin{align*}
        &\mathrm{CrossAttn}(X_{\mathrm{A}},X_{\mathrm{B}})= \\
         &\qquad\mathrm{softmax}(d^{-\frac{1}{2}}(W_{\mathrm{Q}}X_{\mathrm{A}})(W_{\mathrm{K}}X_{\mathrm{B}})^\top)(W_{\mathrm{V}}X_{\mathrm{B}}),
\end{align*}
where $W_Q, W_K, W_V$ are learnable weights.

The input for each layer is given as follows:
\begin{align*}
        \bm{h^{(i)}} = 
        \begin{cases}
            {(\bm{h}_\mathrm{dest}, \bm{h}_\mathrm{obst}, \bm{h}_\mathrm{targ}) \hspace{5mm}} &(i=0)\\
            {(\bm{h}^{(i-1)}_\mathrm{img}, {\bm{h}^{(i-1)}_\mathrm{obst}}) \hspace{5mm}} &(i=1,...,N_I)
        \end{cases}
\end{align*}
Each layer performs the following processing:
\begin{align*}
        \bm{\alpha}^{(i)}_{\mathrm{img}} 
        &= \mathrm{CrossAttn}(\bm{h}^{(i-1)}_\mathrm{img}, \bm{h}_\mathrm{targ}), \\
        \bm{h}^{(i)}_\mathrm{img} &= \mathrm{FFN}(\mathrm{MHA}(\mathrm{FFN}(\bm{\alpha}^{(i)}_\mathrm{img}))),
\end{align*} 
where $\mathrm{FFN}(\cdot)$ denotes the Feed-Forward Network, and $\mathrm{MMHA}(\cdot)$ denotes the Masked Multi-Head Attention\cite{vaswani2017attention}.
$\bm{h}^{(i)}_\mathrm{obst}$ is obtained in a similar manner. Additionally, residual connections and layer normalization were performed after each Feed-Forward Network of the $i$-th $(i = 1, ..., N_I)$ layer.
Finally, the output $\bm{h}_\mathrm{imgs}$ of CAIE is obtained as follows:
\begin{align*}
    \bm{h}_\mathrm{imgs} = (\bm{h}_\mathrm{img}, \bm{h'}_\mathrm{obst}) = (\bm{h}^{(N_I)}_\mathrm{img}, \bm{h}^{(N_I)}_\mathrm{obst}).
\end{align*}

\subsection{Cross Attentional Multimodal Decoder (CAMD)}

This decoder predicts the next token in an auto-regressive manner. The decoder consists of $N_d$ layers. Each layer consists of a Cross-Attention layer and a Feed-Forward Network layer \cite{vaswani2017attention}. The input for each layer is given as follows:
\begin{align*}
        \bm{h}^{(d)}_{\mathrm{dec}} = 
        \begin{cases}
            {(\bm{h}_\mathrm{mul}, \bm{h}_\mathrm{imgs}) \hspace{5mm}} &(i=0)\\
            {(\bm{h}^{(d-1)}_{\mathrm{dec}}, {\bm{h}_\mathrm{imgs}}) \hspace{5mm}} &(i=1,...,N_I)
        \end{cases}
\end{align*}

Given $(\bm{h}_\mathrm{img}, \bm{h}_\mathrm{txt})$, $\bm{h}_\mathrm{mul}$ is obtained as the output of the transformer encoder, which consists of the Masked Multi-Head Attention layer and the Feed-Forward Network layer. Here $\bm{h}_\mathrm{txt}$ denotes the text features. Each layer of the decoder performs the following processing:
\begin{align*}
        \bm{h}^{(d)}_\mathrm{dec} = \mathrm{FFN}(\mathrm{CrossAttn}(\bm{h}^{(d-1)}_\mathrm{dec}, \bm{h}_\mathrm{imgs})).
\end{align*}
Residual connections of $\bm{h}^{(d-1)}_\mathrm{dec}$ and layer normalization were performed after each $i$-th $(i = 1, ..., N_{d})$ layer.
The final prediction probability of the next token, $p(\hat{\bm{y}}_{t+1})$, was calculated using a fully-connected layer and softmax function for output $\bm{h}^{(N_d)}_\mathrm{dec}$ of layer $N_d$-th layer.

\subsection{Nearest Neighbor Captioning Module (NNCM)}
This module enhances the CAMD by employing a nearest neighbors retrieval mechanism.
The input is $(\bm{z}_t, p(\hat{\bm{y}}_{t+1}))$, where $\bm{z}_t$ denotes the latent representation fed into the final Feed-Forward Network layer of the CAMD.

We obtain a set of $N_{\mathrm{knn}}$ pairs $\{(\bm{k}_n, \bm{v}_n)  | n = 1,....,N_{\mathrm{knn}}\}$ from the Datastore\cite{khandelwal2019generalization}, according to the distance function $d(\cdot, \cdot)$ using the k-nearest neighbor method based on $\bm{z}_t$. Here, we used the squared error as $d(\cdot, \cdot)$. The Datastore is defined as follows:
\begin{align*}
        \{(\bm{z}_{i,t}, \hat{\bm{y}}_{i, t+1}) | i=1,...,N, t=1,...,T-1\},
\end{align*}
where $N$ and $T$ denote the number of samples in the training set and the token length of the $i$-th sentence, respectively. In addition, $\hat{\bm{y}}_{i,t+1}$ denotes the $(t+1)$-th token in the $i$-th generated sentence. $\bm{z}_{i,t}$ is obtained from $1$ to $t$-th tokens of the $i$-th reference sentence, $\bm{y}_{i, 1:t}$.
We aggregate the queried $N_{\mathrm{knn}}$ pairs following the equation:
\begin{align*}
        p_{\mathrm{knn}}(\hat{\bm{y}}_{t+1}) = \frac{1}{Z}\bm{V}'\mathrm{softmax}(\bm{k}_{dist}),
\end{align*}
where $\bm{k}_{\mathrm{dist}}$ denotes $\{d(\bm{k}_n, \bm{z}_t)|n=1,...,N_{\mathrm{knn}}\}$ and $Z$ denotes the normalization constant.
$V'$ denotes the stack of the one-hot representation of $\bm{v}_i (i=1,...,N_{\mathrm{knn}})$.
Finally, we obtain the rescored predicted probability  $p_{\mathrm{total}}(\hat{\bm{y}}_{t+1})$, as follows:
\begin{align*}
        p_{\mathrm{total}}(\hat{\bm{y}}_{t+1}) = \lambda_{\mathrm{knn}} p_{\mathrm{knn}}(\hat{\bm{y}}_{t+1}) + (1-\lambda_{\mathrm{knn}})p(\hat{\bm{y}}_{t+1}),
\end{align*}
where $\lambda_{\mathrm{knn}}$ is the weight in the linear interpolation.

\subsection{Loss function}
The loss function is defined as follows:
\begin{align*}
        L = \lambda_{\mathrm{CE}}L_{\mathrm{CE}}(\bm{y}_{t+1}, p(\hat{\bm{y}}_{t+1}))
            + \lambda_{\mathrm{NCE}}L_{\mathrm{NCE}}(\bm{h}_{img}, \bm{h}_{\mathrm{txt}}),
\end{align*}
where $\lambda_{\mathrm{CE}}$ and $\lambda_{\mathrm{NCE}}$ are hyperparameters representing the loss weights.
Additionally, $L_{\mathrm{CE}}$ and $L_{\mathrm{NCE}}$ denote the cross-entropy loss and InfoNCE loss \cite{radford2021learning}, respectively.

\section{Experiments
\label{exp}
}

\vspace{-1mm}
\subsection{Dataset}\label{dataset}
\vspace{-1mm}

\renewcommand{\arraystretch}{1.3}
\begin{table}[t]
    \centering
    \caption{\normalsize Parameter settings}
    \normalsize
    \begin{tabular}{|c|l|}
        \hline
        Loss Weights & $\lambda_{CE}=0.9 , \lambda_{NCE}=5$ \\ \hline
        $N_I, N_d$ & 2 \\ \hline
        $N_{knn}$ & 64 \\ \hline
        $\lambda_{knn}$ & 0.25 \\ \hline
        Epochs &  30 \\ \hline
        Optimizer & Adam$(\beta_1=0.9, \beta_2=0.999)$ \\ \hline
        learning rate &  1.0e-4 \\ \hline
        Batch size & 16 \\ \hline
    \end{tabular}
    \label{tab:parameters}
\end{table}
\renewcommand{\arraystretch}{1.0}

We constructed a dataset using a simulator and an extended version of SIGVerse \cite{inamura2014development}, which was used in the WRS2018 Partner Robot Challenge/Virtual Space Competition \cite{okada2019competitions}.
First, the room layout and the destination were selected randomly. Here, ten different room layouts and six types of destinations were used.
Second, obstacles were placed randomly on the destination. Here, 25 types of obstacles were used.
Third, a target object was randomly selected from 14 types. The robot obtained the RGBD image of the target object and grasped it. The robot placed the objects based on a policy using an attention map for possible collisions obtained from the pre-trained PonNet~\cite{magassouba2021predicting}. According to the policy, the robot places objects in positions with relatively low attention values on the map.
The robot randomly placed the target object in the destination. We then manually extracted scenes where collisions occurred because this study focuses on future captioning related to collisions.
Finally, we asked annotators to give descriptions regarding the collisions.
Note that there is no sufficient space to place the target object in all samples.

We provided the annotators with videos captured from both the robot's camera and a third-party view. They were instructed to annotate the videos with descriptions of any collisions or falls associated with the object placement.

We constructed the BILA-caption 2.0 dataset for the following reasons.
Although CLEVRER \cite{yi2019clevrer} is one of the standard datasets used for generating descriptions of future collisions, it is not suitable for our study.
This is because it is very artificial in terms of robotics and does not assume object placement tasks.
It is problematic when a dataset lacks diversity. Therefore, we have constructed a dataset that is more diverse than CLEVRER.
On the other hand, the BILA-caption dataset \cite{kambara2022relational} is one of the datasets used for future captioning of possible collisions.
However, it is insufficient in our study because it does not contain depth images. 

The size of the BILA-caption 2.0 dataset is 5317 samples. Each sample consists of an RGBD image of a destination, an RGBD image of a target object, and a Japanese description of a collision. The vocabulary size is 1254 and the average sentence length is 22 tokens.
The BILA-caption 2.0 dataset was divided into training, validation, and test sets. The training set, validation set, and test set consisted of 4186, 474, and 657 samples, respectively. Here, we divide the dataset to ensure a consistent proportion of the environment in each set. We used the training set to update the model parameters and the validation set to adjust the hyperparameters. We evaluated our model on the test set.

\begin{table*}[tb]
\centering
\caption{ \normalsize
\textbf{Quantitative comparison and ablation study.}
The best scores are displayed in bold.
}
\begin{tabular}{p{3.2cm}Wc{2.1cm}Wc{2.1cm}Wc{2.1cm}Wc{2.1cm}}\hline
Methods & BLUE-4 & METEOR & ROUGE-L & CIDEr-D  \\ \hline \hline
SAT\cite{xu2015show} & $23.65 \pm 2.00$ & $24.24 \pm 0.96$ 
        & $41.78 \pm 1.58$ & $24.33 \pm 3.56$ \\
RFCM\cite{kambara2022relational} & $25.03 \pm 1.25$ & $27.14 \pm 0.41$ 
        & $44.04 \pm 1.13$ & $25.09 \pm 4.30$ \\
\hline
Ours (w/o NNCM) & $26.39 \pm 0.61$ & $28.63 \pm 0.30$ 
        & $45.54 \pm 0.73$ & $31.34 \pm 1.54$ \\
Ours (w/o CAM) & $26.95 \pm 1.64$ & $28.60 \pm 0.68$ 
        & $45.50 \pm 0.83$ & $32.29 \pm 2.17$ \\
\hline
Ours (full) & $\bm{27.09} \pm \bm{0.70}$ & $\bm{28.82} \pm \bm{0.56}$ 
        & $\bm{46.00} \pm \bm{0.78}$ & $\bm{33.08} \pm \bm{0.73}$ \\
\hline
\end{tabular}
\label{tab:results}
\vspace{-2mm}
\end{table*}

\vspace{-0.5mm}
\subsection{Experimental Settings}
\label{settings}
\vspace{-1mm}

The number of trainable parameters in the proposed method was approximately 190M, and the training was conducted on a hardware configuration consisting of a GeForce RTX 3090 with 24GB of memory and an Intel Core i9 10900K processor, along with 128GB of RAM. It took approximately 1.5 hours and 1 second per sample for model training and inference, respectively.

As a condition for early stopping, we used the generalization \cite{prechelt1998automatic} in $k$-epoch defined by the following equation.
\begin{align*}
        GL(k) = \frac{100 \times E_{va}(k)}{E_{opt}(k)-1}, 
\end{align*}
where $E_{va}(k)$ and $E_{opt}(k)$ denote the verification loss set at the $k$-th epoch and the minimum loss on the verification set up to the $k$-th epoch, respectively.

The CAM was pre-trained using the extended version of the dataset described in \ref{dataset}. 
We used it to train the module according to \cite{magassouba2021predicting}.

\begin{figure}
    \centering
    \includegraphics[clip,width=\linewidth]{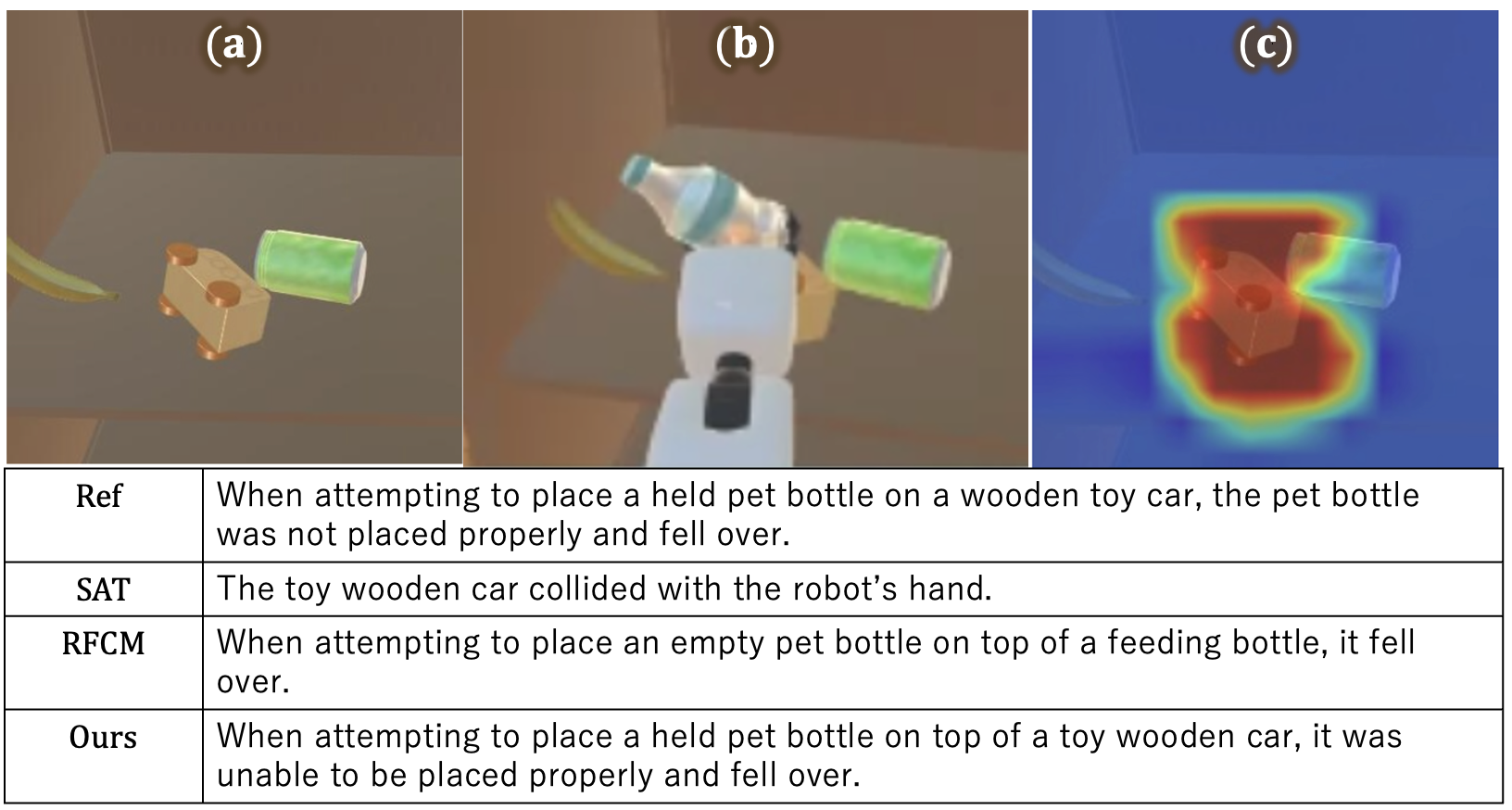}
    \vspace{-6.5mm}
    \caption{\normalsize A successful sample. In this successful sample, the target object is a ``plastic bottle'' and the collided object is a ``wodden toy car.''}
    \label{fig:qual1}
    \vspace{-5mm}
\end{figure}

\begin{figure}
    \centering
    \includegraphics[clip,width=\linewidth]{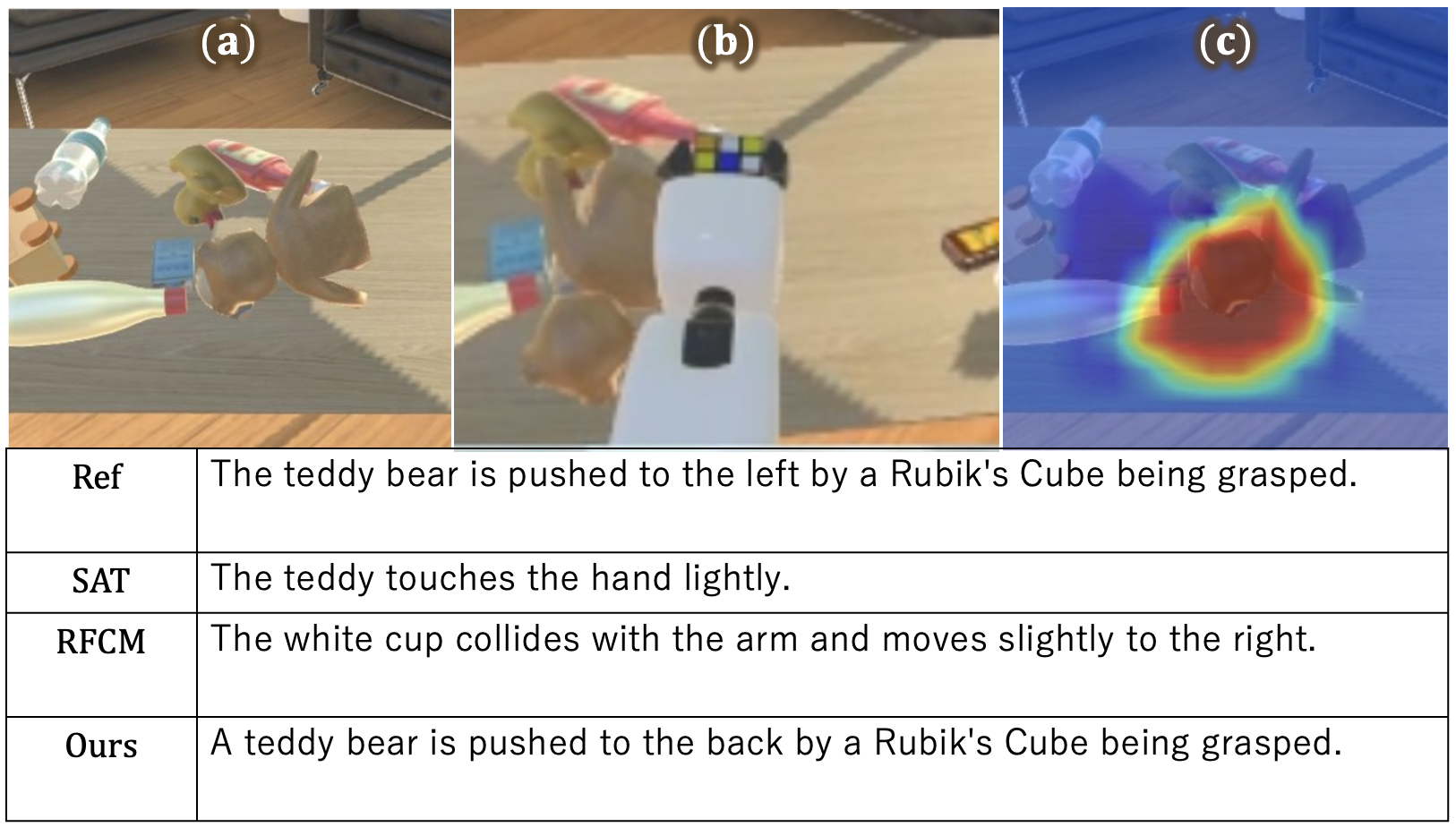}
    \vspace{-8mm}
    \caption{\normalsize A successful sample. In this successful sample, the target object is a ``Rubik's Cube'' and the collided object is a ``teddy bear.''  }
    \label{fig:qual2}
    \vspace{-5mm}
\end{figure}

\vspace{-1mm}
\subsection{Quantitative results}
\vspace{-1mm}

Table~\ref{tab:results} shows the quantitative results for comparing the baseline and proposed methods. Each score represents the mean and standard deviation of five experimental runs.

We selected RFCM \cite{kambara2022relational} as a baseline method because it has been successfully applied to future captioning with respect to collisions in object placement tasks. Additionally, we also selected SAT \cite{xu2015show} because it is a representative image captioning model.

The evaluation of the generated sentences was based on several standard metrics for video captioning tasks: BLEU-4 \cite{papineni2002bleu}, ROUGE-L \cite{lin2004rouge}, METEOR \cite{banerjee2005meteor}, and CIDEr-D \cite{vedantam2015cider}. The primary metric was CIDEr-D. There are multiple possible collisions in the future. It is also challenging to evaluate the task due to the multiple possible solutions. Therefore we selected these metrics following the standard evaluation method of the captioning task.

\begin{figure}
    \centering
    \includegraphics[clip,width=\linewidth]{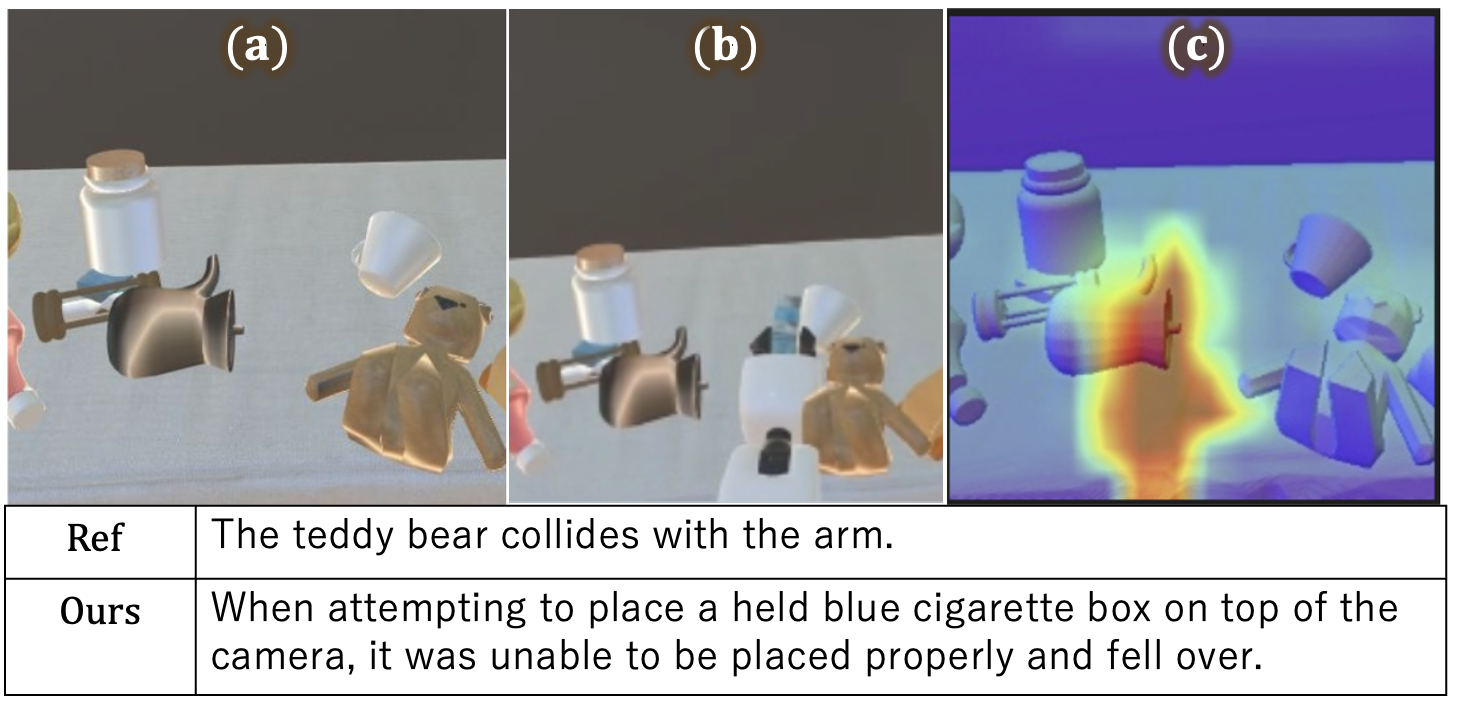}
    \caption{\normalsize A failure sample. In this figure, (a), (b) and (c) show an RGB image of the destination, an image of the moment of collision and an attention map regarding the depth image, respectively.}
    \label{fig:fail}
\end{figure}

Table~\ref{tab:results} shows that our method outperformed the baselines for all metrics. 
Considering CIDEr-D scores, our method achieved 33.08 points, whereas SAT and RFCM achieved 24.33 and 25.09 points, respectively. Therefore, our method outperformed SAT and RFCM by 8.75 points and 7.99 points, respectively. 
Similarly, for BLEU-4, METEOR, and ROUGE-L, our method outperformed SAT and RFCM by 3.44 and 2.06 points, 4.58 and 1.68 points, and 4.22 and 1.96 points, respectively. The difference in performance between the proposed method and baseline methods was statistically significant ($p<0.05$) for all standard evaluation metrics.

\vspace{-1mm}
\subsection{Qualitative results}
\vspace{-1mm}

Figs.~\ref{fig:qual1} and~\ref{fig:qual2} show successful examples. 
In these figures, (a), (b) and (c) show an RGB image of the destination, an image of the moment of collision and an image obtained by overlaying the attention map on the RGB image of the destination, respectively.

In Fig~\ref{fig:qual1}, the target object is a ``plastic bottle'' and the collided object is a ``wooden toy car.''
SAT incorrectly described the collided object as the ``hand'' instead of a ``woden toy car''
Similarly, RFCM incorrectly described the collided object as a ``baby bottle''
In contrast, our model appropriately described the target object and the collided object as a ``plastic bottle'' and ``wooden toy car,'' respectively.
In addition, in Fig.~\ref{fig:qual2}, the target object is a ``Rubik's Cube'' and the collided object is a ``teddy bear.''
SAT incorrectly described the collided object as the ``hand'' instead of a ``Rubik's Cube.'' RFCM also incorrectly described the collision between the ``arm'' and a ``white cup.''
In contrast, our method appropriately described the collision between the ``Rubik's cube'' and the ``teddy bear.''

Fig.~\ref{fig:fail} shows a failure sample. In this figure, (a), (b) and (c) show an RGB image of the destination, an image of the moment of collision and an attention map regarding the depth image, respectively.
In the example shown in Fig.~\ref{fig:fail}, the reference sentence was ``The arm collides with the teddy bear.'' Our method incorrectly generated ``When attempting to place the blue box of cigarettes onto the camera, it results in the cigarette box falling over.''
In the sentence, multiple objects related to the collision were inappropriately described.
It is considered that the failure is attributed to the wrong attention map.
Indeed, the right figure shows that attention is not focused on the teddy bear, but rather on the black soy sauce bottle. 
This indicates that the black soy sauce bottle is mistaken for a camera with a similar color.

\vspace{-2.0mm}
\subsection{Ablation Study}
\vspace{-1mm}

Table~\ref{tab:results} shows the quantitative results of the ablation study. 
We defined the following two ablation conditions.
\begin{enumerate}[label=(\roman*)]
    \item \textit{NNCM Ablation}: 
          We removed the NNCM to investigate its contribution to performance.
          Table~\ref{tab:results} shows that the CIDEr-D score was 31.34, which was a decrease of 1.74 points compared with our full model.  In particular, for the primary metric, CIDEr-D, there was a significant difference in performance between ``Ours (full)'' and ``Ours (w/o NNCM)'' ($p < 0.05$). This indicates that the model acquired the capability to capture and generate the diverse expressions presented in the training set.
    \item \textit{CAM Ablation}:  
          We removed the CAM to investigate its contribution to performance.
          Table~\ref{tab:results} shows that the CIDEr-D score was 32.29, which was a decrease of 0.79 points compared with our full model. 
          The results of (ii) show the effectiveness of the CAM. The CAM appropriately focused attention on the regions of potential collision. This enabled the precise generation of descriptions regarding collision objects as shown in the qualitative results in Fig. \ref{fig:qual1} and \ref{fig:qual2}, leading to improved performance.
\end{enumerate}

These results indicate that introducing the retrieval mechanism in the NNCM significantly influenced the performance.

\vspace{-1mm}
\subsection{Subject Experiment}
\vspace{-1mm}

We conducted a subject experiment and used the Mean Opinion Score as a metric. In the experiment, 100 episodes were randomly selected from the test set. For each episode, we generated captions using SAT \cite{xu2015show}, RFCM \cite{kambara2022relational}, and the proposed method. In the experiments, we evaluated a total of 400 sentences, consisting of 300 sentences from the generated captions and 100 reference (ground truth) sentences.
Eight subjects participated in the experiment. We presented 50 sentences and their corresponding episodes to each subject and asked for their evaluation. The subjects evaluated the sentences in terms of their intelligibility using a 5-point scale:

1, Very bad; 2, Bad; 3, Normal; 4, Good; 5, Very good.

Table~\ref{tab:mos_comparison} shows the quantitative results. Each score represents the mean and standard deviation. 
Our method achieved 2.47 points, whereas SAT and RFCM achieved 1.89 and 2.15 points, respectively. Therefore, our method outperformed SAT and RFCM by 0.58 and 0.32 points, respectively. This result demonstrates that the quality of the sentences generated using the proposed method is higher than that using the baselines, based on human evaluation.

\begin{table}[h]
\centering
\caption{\normalsize Comparison of methods based on Mean Opinion Score}
\label{tab:mos_comparison}
\begin{tabular}{@{}lc@{}}
\toprule
Method & Mean Opinion Score \\ 
\midrule
SAT    & $1.89 \pm 1.07$ \\
RFCM   & $2.15 \pm 1.15$ \\
Ours   & $\bm{2.47 \pm 1.22}$ \\
Reference & $4.34 \pm 0.91$ \\
\bottomrule
\end{tabular}
\end{table}

\vspace{-5mm}
\subsection{Error Analysis}
\vspace{-1mm}


\begin{table}[t]
\centering
\caption{\normalsize  An error analysis of failure cases.}
\label{table:errors}
\renewcommand{\arraystretch}{1.3}
\tiny
\begin{adjustbox}{width=\linewidth}
\begin{tabular}{clc}
\hline
Error ID  & \multicolumn{1}{c}{Description} & \# Samples \\ \hline
ODE    & Obstacle Description Error    & 31   \\
SE    & Serious description error regarding collisions   & 8   \\
OUG   & Over / Under generation               & 7   \\
Others   & Others                   & 4   \\
\hline
Total & -                                & 50  \\ \hline
\end{tabular}
\end{adjustbox}
\end{table}
\renewcommand{\arraystretch}{1.0}



We analyzed errors for 50 samples with the lowest CIDEr-D scores. Table \ref{table:errors} categorizes the failure cases. The causes of failure can be roughly divided into four groups:

\begin{enumerate}[label=(\roman*)]
    \item Obstacle Description Error (ODE): \mbox{}\\
          This category refers to failure cases, including the incorrect description of a single object. For example, in a sample where a plastic bottle collides with a black soy sauce container, our model generated ``A plastic bottle collides with a teddy bear''.
    \item Serious description error regarding collisions (SE):  \mbox{}\\
          This category contains the cases where the model generated incorrect sentences that wrongly describe potential collisions or include more than two errors regarding obstacles associated with collisions.
          For example, in a sample where an hourglass collided with a salt container, which resulted in rolling a plastic bottle, our model inappropriately generated ``the arm collides with the teddy bear''.
    \item Over/Under generation (OUG):  \mbox{}\\
          This category includes the cases where the generated sentences did not entirely depict the collision, or they described events that did not occur.
          For example, in a sample where a container filled with salt falls from the furniture where it was placed because of a collision, our model generated ``The container filled with salt collided'' without mentioning the fact that it fell from the furniture because of a collision.
    \item Others \mbox{} \\ 
\vspace{-1.0mm}
\end{enumerate}

Table~\ref{table:errors} shows that the main cause of errors is ODE. This indicates that the major error involves the failure to identify the objects associated with the collisions. 
In the CAMD, the latent representation of language is always used as the query and the latent representation of the image is always used as the key and value. Potentially, only the fixed relationship between the language and image was modeled, which resulted in the insufficient representation of the image.
As an improvement, for instance, introducing an attention mechanism with symmetry could be considered.

\vspace{-1.0mm}
\section{Conclusions}

In this paper, we handled future captioning tasks for generating descriptions of situations given past information.
In particular, we focused on future captioning related to collisions that occur when a DSR places an object on a piece of furniture that has obstacles on it.
We would like to emphasize the following contributions of our study:
\begin{itemize}
 \item We proposed NNFCM that introduces NNLM for the future captioning of possible collisions. This is one of the first attempts to introduce NNLM into multimodal language generation.
 \item  We introduced the Collision Attention Module, which extracts attention regions for possible collisions.
 \item We introduced the Cross Attentinal Image Encoder, which models the relationship between the target object and the destination.
 \item Our method outperformed the baseline methods on the BILA-caption 2.0 dataset.
\end{itemize}

In future work, we plan to extend our model to simultaneously predict collisions and generate descriptions.

\section*{ACKNOWLEDGMENT}
This work was partially supported by JSPS KAKENHI Grant Number 20H04269, JST Moonshot, JST CREST, and NEDO.

\bibliographystyle{tfnlm}
\bibliography{reference}
\end{document}